\documentclass{JMLFS}

\journal{JMLFS-ID}
\vol{2022}

\received{4 January 2022}
\published{xx Month 2022}

\def\be{\begin{equation}}
\def\ee{\end{equation}}
\def\bea{\begin{eqnarray}}
\def\eea{\end{eqnarray}}

\usepackage{listings}
\usepackage[section]{placeins}

\begin{document}

\title{Applications of Gaussian Mutation for Self Adaptation in Evolutionary Genetic Algorithms}

\author{Okezue Bell\auno{1}$^,$\auno{2}}
\address{$^1$Harvard Medical School-MIT, aobell@mgh.harvard.edu}
\address{$^2$Moravian Academy Upper School}

\begin{abstract}
In recent years, optimization problems have become increasingly more prevalent due to the need for more powerful computational methods. With the more recent advent of technology such as artificial intelligence, new metaheuristics are needed that enhance the capabilities of classical algorithms. More recently, researchers have been looking at Charles Darwin's theory of natural selection and evolution as a means of enhancing current approaches using machine learning. In 1960, the first genetic algorithm was developed by John H. Holland and his students (Holland, 1975). We explore the mathematical intuition and implications of the genetic algorithm in developing systems capable of evolving using Gaussian mutation.
\end{abstract}

\maketitle


\section{Gaussian Mutations in Genetic Algorithms}

\begin{itemize}
\item {\bf Introduction to genetic algorithms}
\item {\bf Self-adaptation}
\item {\bf The Gaussian mutation operator}
\item {\bf Implementation of gaussian mutation results}
\item {\bf Conclusion}
\end{itemize}

\section{{Introduction to genetic algorithms}}
A genetic algorithm is a type of stochastic search algorithm that functions off of the fundamental principles of natural selection. Essentially, the algorithm works off of given string structures, typically organisms, which gradually evolve, and their fitness is characterized by survival and reproduction. The genetic algorithm sustains the survival of the fittest rule, where there is a randomized information exchange but with a structured method (Gholizadeh, 2013). With every succeeding generation, new strings are created and the fittest strings from the previous generations are used as a reference. The "winners" are considered the individuals whose genes enable them to survive and reproduce. With each successive selection, more resources are retained. The algorithm is capable of simulating this process and uses it to calculate the optimum of objective functions, thus making the algorithm suitable for solving optimization problems. In summary, the genetic algorithm (GA) uses neuro-evolutionary concepts and artificial intelligence to design better neural networks (Roetzel et al., 2020).

When most of the algorithm's intricacies are removed, it essentially takes in a set of something and evolves it to its optimal state. Figure 1 shows the abbreviated process for a cluster of carbon atoms being evolved to organize in the lowest energy carbon structure (Figure 1).

\begin{figure}[ht]
\centering
\includegraphics[height=3in]{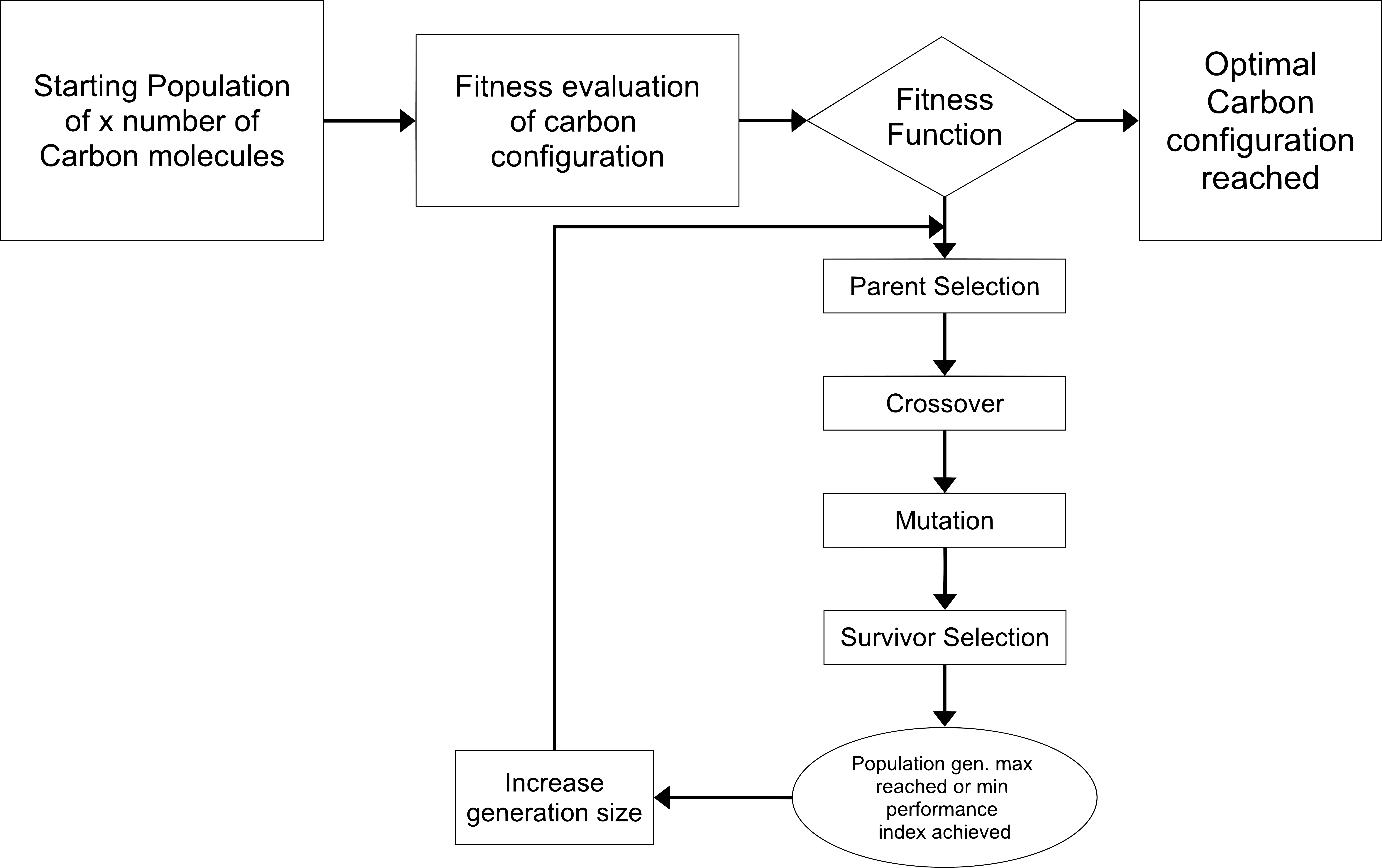}
\caption{The GAs process of determining the buckminsterfullerene as the lowest energy configuration of 60 carbon atoms. It has a carbon population and knows it wants to solve for the lowest energy structure. It then generates a bunch of random configurations, which are then analyzed for their fitness based on their energy levels. The best ones mate and create a new generation, and this process repeats until an ultralow energy level generation is created (the buckminsterfullerene structure). It took scientists over 3 years to solve this, while the GA can complete this process in about 4 hours (self-made).}
\end{figure}

GAs have been applied to a variety of fields, including inverse kinematic problems (IKPs) for advanced robotic designs for transhumeral prosthetics (Števo et al., 2019) and tumor tracking in combination with intermittent Kalman predictors (IKPs) by using genetic algorithms to solve combinatorial optimization problems to conclude the measurements of tumor motion (Aspeel et al., 2019). Currently, the popular usage of GAs falls in the category of generating an ANN capable of determining a very precise solution for an infinite set of possibilities, more specifically a search or optimization problem. Genetic algorithms are considered a subset of artificial intelligence and are a part of the broader scope of evolutionary computing (Eiben, 2003).

\section{{ Self-adaptation}}
What is currently considered one of the most exciting fields of research for evolutionary computing and genetic algorithms (Hinterding et al., 2005) is the concept of self-adaptation in GAs, especially for numeric functions. Self-adaptation is the property of a genetic algorithm to adapt its algorithm to a specific problem when calculating the solution for it. This can be done for one or more facets of evolution, and approaches for self-adapting for both mutation strength and population size, for example, have been successful on initial testing (Hinterding et al., 2005). The mutation genetic operator introduces genetic diversity with the genetic algorithm population of chromosomes. It can be likened to biological mutation. There are many types of mutation operators, including bit string, boundary, and Gaussian, most of which deal with the introduction of a random value to a specific gene, which influences the expression of the offspring.

The Gaussian mutation operator has proven to be an optimal and popular choice for self-adaptation in genetic algorithms (Li-hui Zhang Ph.D. et all., 2015). In Gaussian mutation, a random unit Gaussian distribution value is added to a selected gene by the operator. The value is thus added to each element of the gene carrier's vector, which causes the development of a new offspring (Figure 2). 
\begin{figure}[ht]
\centering
\includegraphics[height=0.25in]{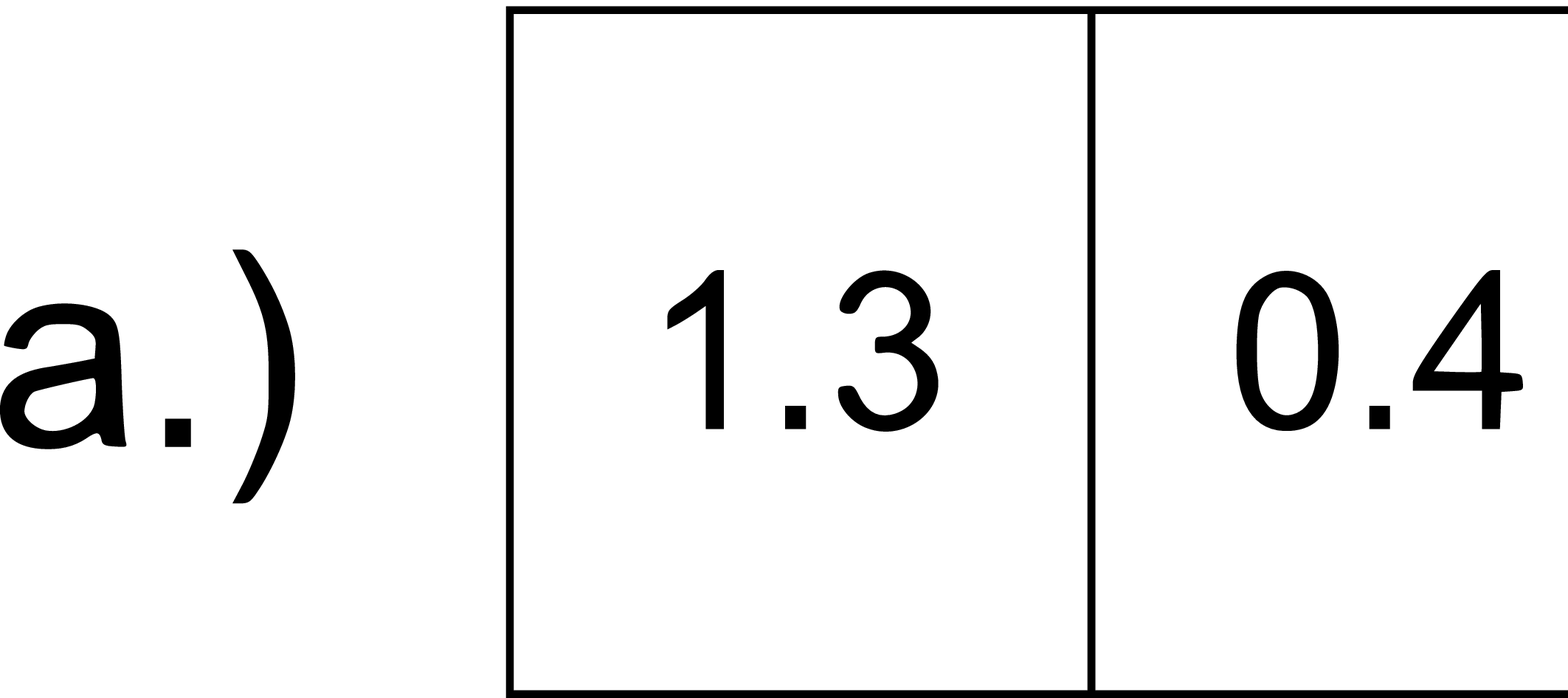}
\caption{Gaussian Mutation of the a) parent to form the b) offspring, causing variance (Dorronsoro, 2004).}
\end{figure} Aside from Gaussian mutation, there are other popular operators, including the one-point crossover reproduction operator (Figure 3), and the bit-flipping mutation operator (Figure 4).
\begin{figure}[ht]
\centering
\includegraphics[height=1in]{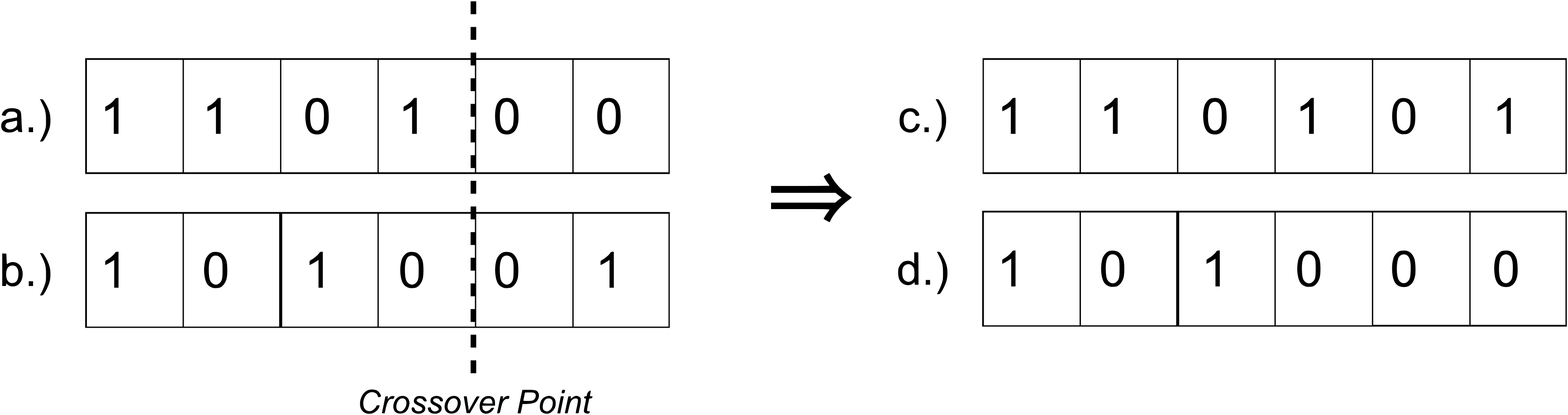}
\caption{One point crossover of parents a) and b) to form the c) and d) offspring from reproduction. A subsequence of the parents, represented as strings, is swapped to create two new offspring (Dorronsoro, 2005).
}
\

\centering
\includegraphics[height=0.25in]{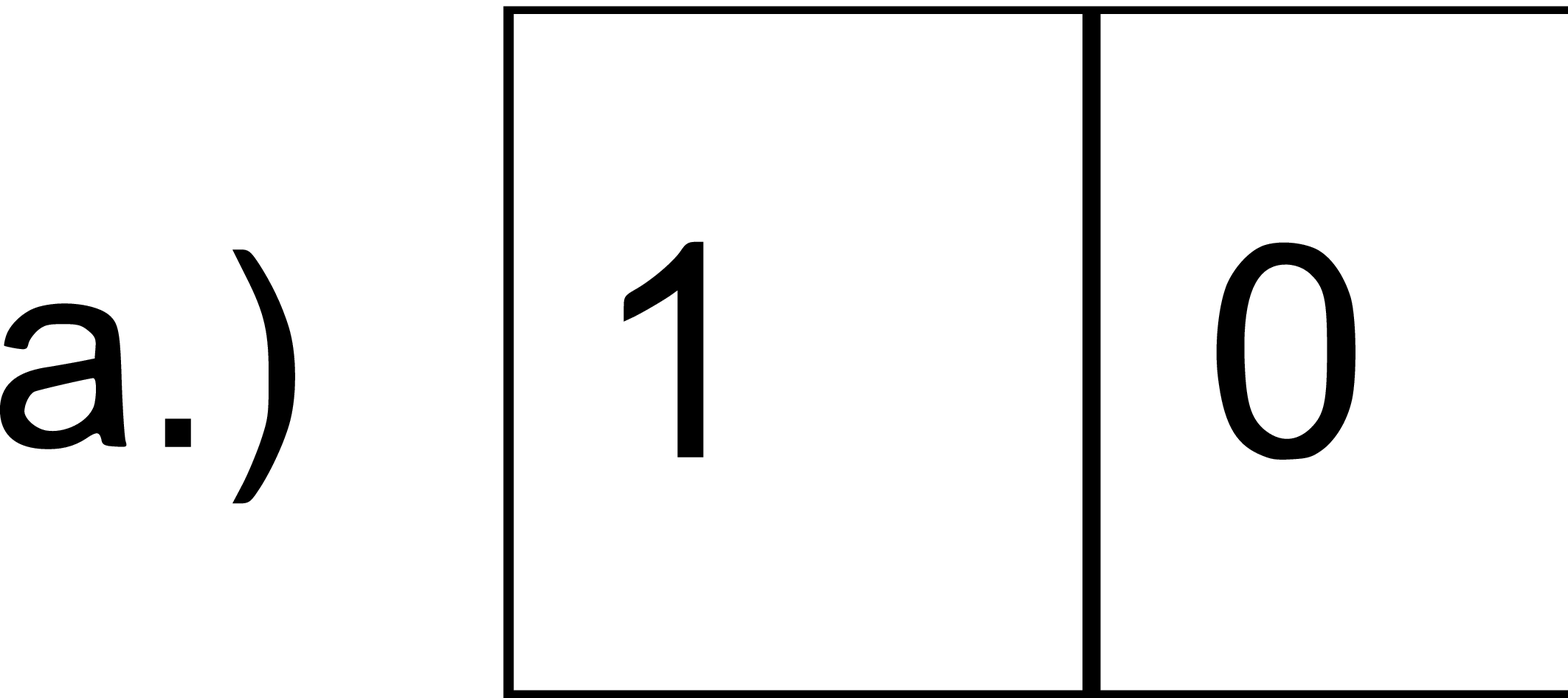}
\caption{The bit flipping mutation of the a) parent to form the b) offspring. A singular binary bit in the original parent string is flipped to create a new string for a different offspring (Dorronsoro, 2005).}
\end{figure}
However, Gaussian mutation operators are still widely used in genetic algorithms, especially in a variety of object-oriented programming languages and their use of Gaussian distribution values. This is because of the Gaussian curve's (Figure 5) usage in probability density functions and standard deviation. Usage of Gaussian values is promising for mutation simulations, and when introducing new information to the population of the GA.
\begin{figure}
\centering
\includegraphics[height=2in]{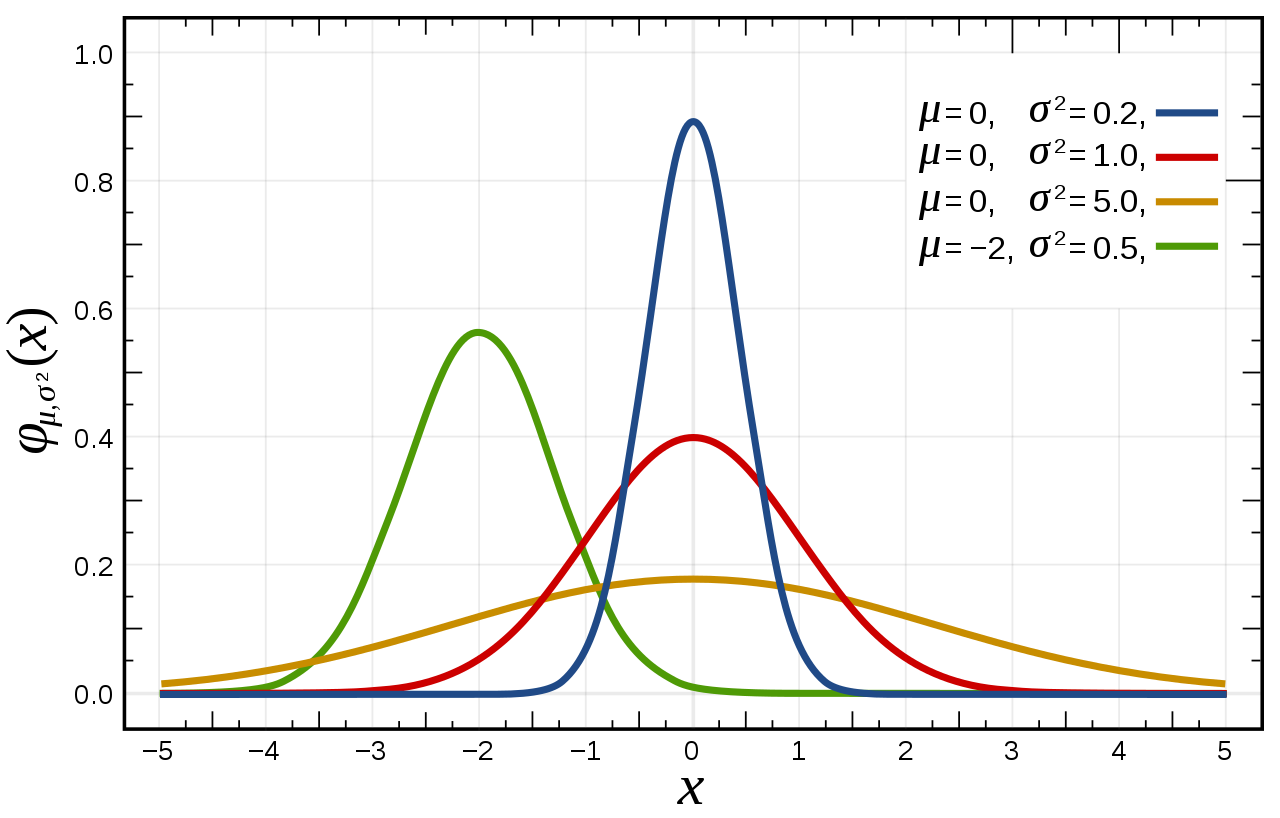}
\caption{A selection of Normal Distribution Probability Density Functions (PDFs). Both the mean, $\mu$, and variance, $\sigma^2$, are varied. The red curve on the graph represents the standard normal distribution. The key is given on the graph (Inductive Load, 2008).}
\end{figure}

\section{{The Gaussian mutation operator}}
Take $x \in [a, b]$ as a real variable. The Gaussian mutation operator $M_g$ changes the variable $x$ to

\begin{center} $M_g(x) := min(max(N(x, \sigma),a),b)$ \end{center}

\noindent The operator also leverages the Gauss error function. Recall that all solutions in the genetic algorithms are the chromosomes. These chromosomes go through mutation (random gene alterations), crossover (swapping genetic material), and selection (only the fittest survive). Mutation operators use a single parent chromosome and induce mutation on some of the selected genes. The offspring when using the Gaussian mutation operator is given by

\begin{center} $x^{'}_{i} = \sqrt{2} \sigma (b_i - a_i)erf^{-1}(u^{'}i)$  \end{center}

\noindent , where  $x^{'}_{i}$ is the offspring, $\sigma$ is a fixed parameter for all variables, and $[a_i, b_i]$ represents the range of random gene $x_i$. $erf^{-1}$ represents the inverse of $erf$, which is the Gauss error function, defined by $erf(y) = \frac{2}{\sqrt{\pi}}\int_{0}^{y} e^{-t^{2}}\,dt$. The inverse Gauss error function is applied to a calculation ui,, which is a random value $u_i$  within the range of $(0,1)$, which is inside of a formula that gives $u_i$, ("Mutation Algorithms for Real-Valued Parameters (GA)", 2018). This formula can be defined using IFTTT (if this then that) notation.

With evolutionary programming, the manipulation of parameters is done through self-adaptation. This learning format allows the GA to constantly tweak the strategy parameters when it's iterating through different generations to find the optimum, essentially searching for the value of the parameters themselves to increase their performance levels. Parametric optimization values are subject to change; their fluctuations depend on the solved function. With Gaussian mutation, convergence is completed much more efficiently, and domain searching and solution tuning are much better (Hartfield, 2010), thereby increasing the performance of the GA.

The Gaussian mutation's shape can also be controlled by applying a parameter to it. This mutation operator is called the $\emph{q}$-Gaussian mutation, where the shape of the mutation distribution is controlled via a real parameter $\emph{q}$ (Tinós et all., 2011). The $\emph{q}$-Gaussian, therefore, operates on a different distribution than the traditional Gaussian mutation. This is known as the q-analog of the normal curve, and its distribution is symmetrical at about 0. Figure 6 shows its probability density function.

\begin{figure}[ht]
\centering
\includegraphics[height=2.5in]{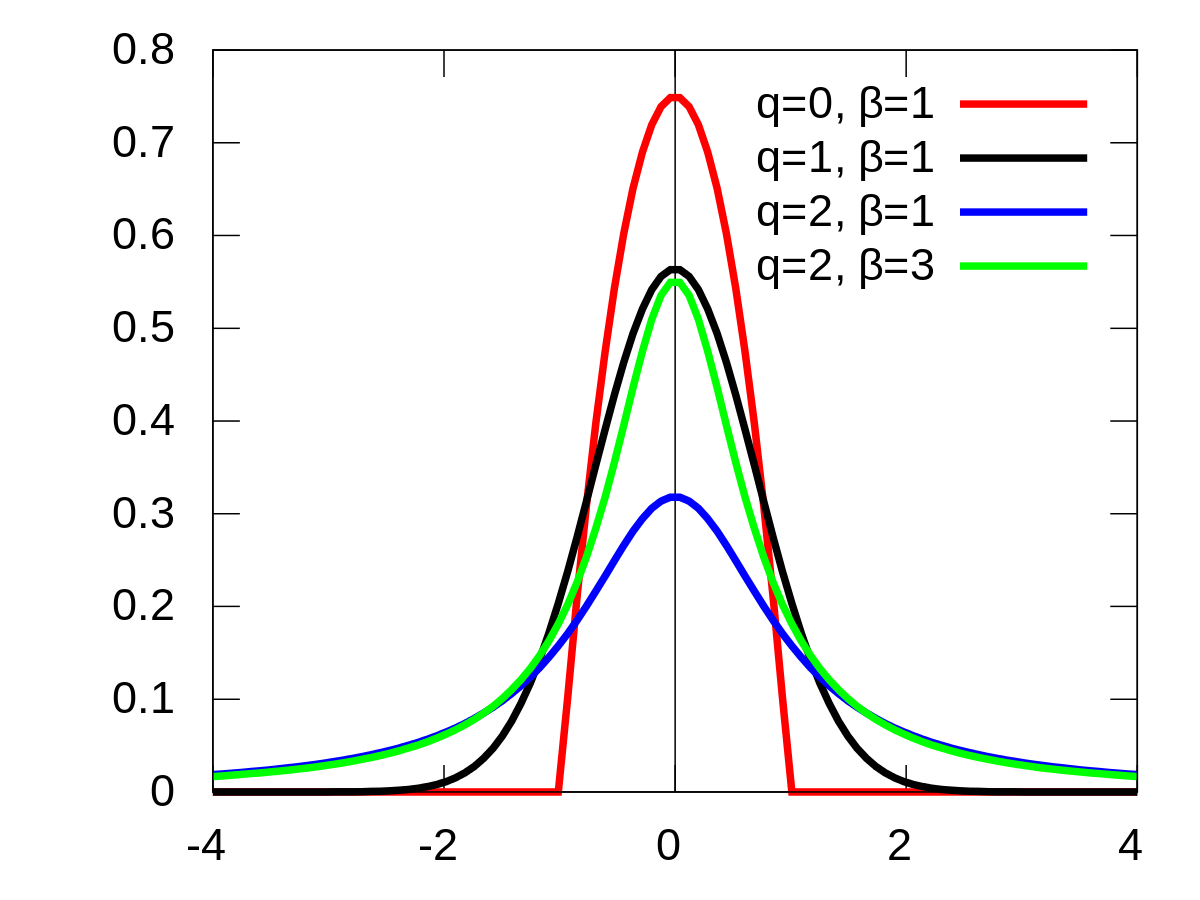}
\caption{Probability density plots of $\emph{q}$-Gaussian distributions. This maps the probability density function of the q-Gaussian distribution itself (IkamusumeFan, 2014).}
\end{figure}

$\emph{q}$-Gaussians allow the mutation distribution shape of the GA self-adaptive, which makes them effective in solving dynamic optimization problems (Yang et al., 2010), where the variable values are subject to change over time.

\section{Implementation of gaussian mutation results}
The implementation of Gaussian mutation has been successful in solving a variety of optimization problem types. A notable instance of this is in $\emph {An Improved Real-Coded Genetic Algorithm Using the Heuristical Normal Distribution and Direction-Based Crossover}$, where a multi-offspring improved real-coded genetic algorithm (MOIRCGA) using the Gaussian distribution (in combination with direction-based crossover was proven to solve constrained optimization problems. The MOIRCGA finished with supremacy over a real-coded genetic algorithm (RCGA) not leveraging Gaussian distribution to find a globally optimal solution to the problem (Wang et al., 2019), in part due to its much faster conversion speed than the RCGA (Figure 7).
\begin{figure}[ht]
\centering
\includegraphics[height=3.8in]{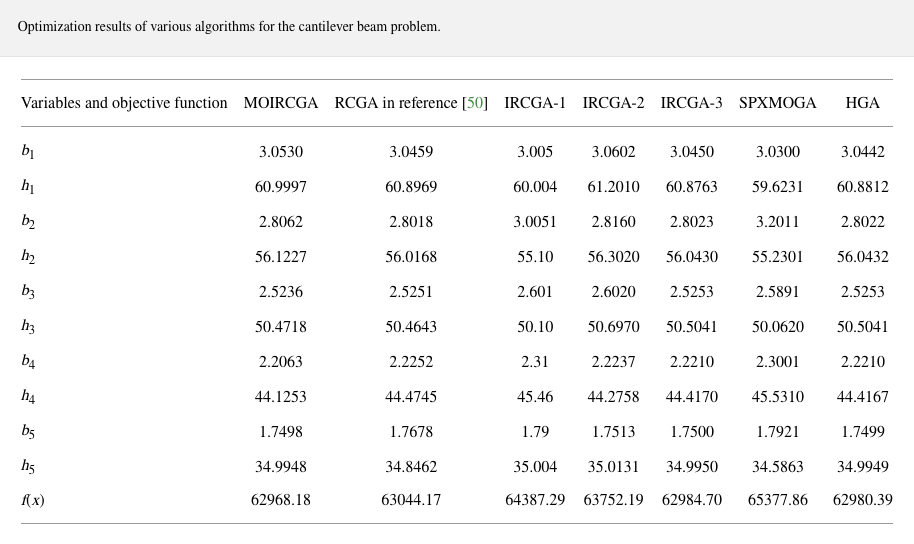}
\caption{Optimization results for the MOIRCGA as compared to the RCGA in the  An Improved Real-Coded Genetic Algorithm Using the Heuristical Normal Distribution and Direction-Based Crossover Experiment (Wang et al., 2019). The MOIRCGA was superior to the RCGA and some of the other benchmarked algorithms.}
\end{figure} 

\noindent Within this experiment, the algorithms were tested and compared with other literature, and the problem was used in the parameter optimization of the Cantilever Beam design with a discrete cross-sectional area to determine the validity of the MOIRGCA, and sixteen test functions were evaluated as well. The experimentation concludes with, "optimization results show that the function value obtained with MOIRCGA is superior to that obtained with RCGA'' (Wang et al., 2019).

Within this experiment, the algorithms were tested and compared with other literature, and the problem was used in the parameter optimization of the Cantilever Beam design with a discrete cross-sectional area to determine the validity of the MOIRGCA, and sixteen test functions were evaluated as well. The experimentation concludes with, "optimization results show that the function value obtained with MOIRCGA is superior to that obtained with RCGA'' (Wang et al., 2019).

\subsection {Robotic appendages.} Genetic algorithms have been cited as one of the most useful forms of artificial intelligence (Cheng, 2011). The algorithms have extensive implications in robotics, especially within movement or functionality (Davidor, 1991). The robotics inverse kinematics problem (IKP) describes "finding a vector of joint variables which produce [the] desired end-effector location" (DeMers et al., 1997). The problem has an infinite amount of solutions, making it difficult to solve using classical computation. However, genetic algorithms have been presented to create the best robotic arm (Sekaj et al., 2014) based on trajectory control. In $\emph {Optimization of Robotic Arm Trajectory Using Genetic Algorithm}$, a genetic algorithm that could optimize for energy consumption, operating time, rotation changes, and trajectory was proposed for the industrial robot ABB IRB 6400FHD (Števo et all., 2014). In $\emph {Genetic Algorithm Based Approach for Autonomous Mobile Robot Path Planning}$, a GA with an improved crossover operator was suggested to be applied to a path planning problem for a computer vision autonomous robot for movement and navigation.
\subsection {Molecular geometry.} In the introductory section of this independent study, an example of determining carbon fullerenes as the lowest energy configuration of sixty carbon atom clusters. This example for a genetic algorithm was extracted from $\emph{Molecular Geometry Optimization with a Genetic Algorithm}$, in which a genetic algorithm was used to determine the lowest energy geometry for carbon compounds, ranging up to $C_60$. The experiment, however, wasn't limited, as their method was capable of determining the lowest energy structure of an atomic cluster in any arbitrary model potential. Original solution candidates generated were used as a reference for fitness to generate a new generation of solutions to produce lower energy solutions. The genetic model itself was cited to be more accurate than the simulated annealing, and was stated to have "dramatically outperformed" it, and decreased the computational time of concluding carbon fullerenes by over 200$\%$ when compared to a human biochemist (Deaven et al., 1995). In $\emph{First-Principles Molecular Structure Search with a Genetic Algorithm}$, a genetic algorithm was built using Conda's RDKit library in Python to accelerate computational chemistry. The algorithm focused on identifying low-energy conformers for a given molecule using first principle searches (Supady et al., 2015). The algorithm was successful (Figure 8) in doing so.\begin{figure}[ht]
\centering
\includegraphics[height=3in]{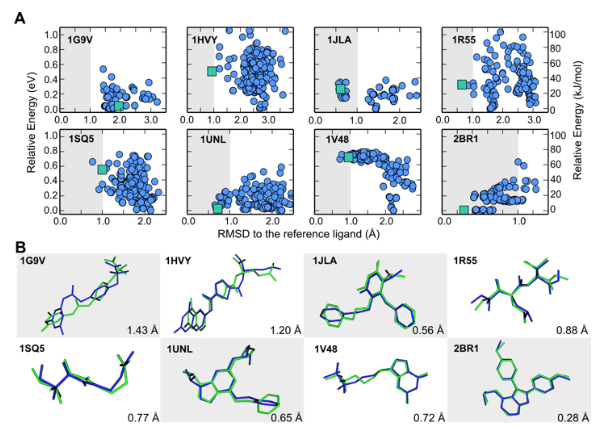}
\caption{"Evaluation of the results for the subset of the Astex Diverse Set. (A) The relative energy of all found conformers as a function of the RMSD to the reference ligand (blue circles). The green squares depict the reference ligand structures after DFT optimization. (B) An overlay between the reference ligand (green) and the best match (blue) is presented together with the corresponding RMSD value" (Supady et al., 2015).}
\end{figure}\subsection{Sentiment Lexicon Optimization.} Another use of genetic algorithms is in analyzing emotions, also known as opinion mining or sentiment analysis. In ALGA: Adaptive lexicon learning using genetic algorithm for sentiment analysis of microblogs, the common problem in sentiment analysis, "improving polarity classification of sentiments in microblogs by building adaptive sentiment lexicons'' (Keshvarz et al., 2017), was solved using a novel genetic algorithm. In the six datasets through which it was tested, the algorithm achieved over 80$\%$ accuracy. In summary, the algorithm was able to build the most accurate collection of words, and how their semantics relate to their sentiment orientation (Keshvarz et al., 2017).

\section{Conclusion}

\section*{Acknowledgements}
This research was conducted independently and presented at a UPenn SEAS functional computing paper competition. Author Okezue Bell is affiliated with both Harvard-MIT and Moravian Academy, though this research was conducted independently.

\bibliographystyle{unsrt}

\begin{thebibliography}{99}
\bibitem{mr01} Keshavarz H, Abadeh MS (2017) ALGA: Adaptive lexicon learning using genetic algorithm for sentiment analysis of microblogs. Retrieved December 19, 2021, from https://doi.org/10.1016/j.knosys.2017.01.028.
\bibitem{m04} Deaven DM, Ho KM (1995) Molecular Geometry Optimization with a Genetic Algorithm. In: Physical Review Letters.  Retrieved December 19, 2021, from https://doi.org/10.1103/PhysRevLett.75.288.
\bibitem{SNO} Lamini C, Benhlima S, Elbekri A (2018) Genetic Algorithm Based Approach for Autonomous Mobile Robot Path Planning. In: Procedia Computer Science. Retrieved December 19, 2021, from https://doi.org/10.1016/j.procs.2018.01.113.
\bibitem{ikoost10} Inverse Kinematics Problem. In: Inverse Kinematics Problem - an overview | ScienceDirect Topics. Retrieved December 19, 2021, from https://www.sciencedirect.com/topics/engineering/inverse-kinematics-problem. 
\bibitem{af10} Števo S, Sekaj I, Dekan M (2016) Optimization of Robotic Arm Trajectory Using Genetic Algorithm. In: IFAC Proceedings Volumes. Retrieved December 19, 2021, from https://doi.org/10.3182/20140824-6-ZA-1003.01073.
\bibitem{af05} Genetic Algorithms and Robotics. In: World Scientific. Retrieved December 19, 2021, from https://doi.org/10.1142/1111.
\bibitem{bh05} VII Jornadas Zaragoza-Pau de Matemática Aplicada y estadística : Jaca (Huesca), 17-18 de septiembre de 2001, 2003-01-01, ISBN 84-96214-04-4, pags. 27-36.
\bibitem{c78} CSDL: IEEE Computer Society. In: CSDL | IEEE Computer Society.  Retrieved December 19, 2021, from https://doi.org/10.1109/SBRN.2010.46.
\bibitem{w78} R. Tins and S. Yang, "Evolution Strategies with q-Gaussian Mutation for Dynamic Optimization Problems," in Neural Networks, Brazilian Symposium on, Sao Bernardo do Campo, Sao Paulo Brazil, 2010 pp. 223-228. Retrieved December 19, 2021, from https://doi.org/10.1109/SBRN.2010.46.
\bibitem{m10-1} Pontes EAS (2018) A Brief Historical Overview Of the Gaussian Curve: From Abraham De Moivre to Johann Carl Friedrich Gauss. In: ijesl.org. Retrieved December 19, 2021, from http://www.ijesi.org/papers/Vol(7)i6/Version-5/D0706052834.pdf.
\bibitem{m06-1} Hinterding R, Michalewicz Z, Eiben AE (2005) Adaptation in Evolutionary Computation: A Survey. Retrieved December 19, 2021, from https://cs.adelaide.edu.au/~zbyszek/Papers/p34.pdf.
\bibitem{m06-1} Aspeel A, Dasnoy D, Jungers Raphael M, Macq B (2019) Optimal Intermittent Measurements for Tumor Tracking in X-ray Guided Radiotherapy. Retrieved December 19, 2021, from  https://arxiv.org/pdf/1903.08990.pdf.
\end{thebibliography}

\end{document}